\documentclass[a4paper]{article}

\usepackage{INTERSPEECH2020}
\usepackage{multirow}
\usepackage{multicol}
\usepackage{cite}

\title{Self-supervised Pre-training with Acoustic Configurations \\for Replay Spoofing Detection}
\name{Hye-jin Shim$^1$, Hee-Soo Heo$^2$, Jee-weon Jung$^1$, and Ha-Jin Yu$^{1\dag}$\thanks{$^\dag$ Corresponding author}}
\address{$^1$School of Computer Science, University of Seoul, Republic of Korea\\
$^2$Naver Corporation, Republic of Korea}
\email{shimhz6.6@gmail.com, heesoo.heo@navercorp.com, jeewon.leo.jung@gmail.com, hjyu@uos.ac.kr}

\begin{document}

\maketitle
\begin{abstract}
Constructing a dataset for replay spoofing detection requires a physical process of playing an utterance and re-recording it, presenting a challenge to the collection of large-scale datasets. 
In this study, we propose a self-supervised framework for pre-training acoustic configurations using datasets published for other tasks, such as speaker verification. Here, acoustic configurations refer to the environmental factors generated during the process of voice recording but not the voice itself, including microphone types, place and ambient noise levels.
Specifically, we select pairs of segments from utterances and train deep neural networks to determine whether the acoustic configurations of the two segments are identical.
We validate the effectiveness of the proposed method based on the ASVspoof 2019 physical access dataset utilizing two well-performing systems.
The experimental results demonstrate that the proposed method outperforms the baseline approach by 30\%.
\end{abstract}

\section{Introduction}
Recent advances in audio spoofing techniques, such as voice conversion, speech synthesis, and replay attacks, threaten the reliability of automatic speaker verification (ASV) systems.
To protect ASV systems against spoofing attacks, the initiative called ASVspoof has been led the research on audio spoofing detection by collecting and sharing common datasets, as well as holding competitions to facilitate thorough investigation \cite{wu2015asvspoof, Kinnunen2017, Todisco2019}. 
ASVspoof 2019 aims to develop countermeasures for three major attack types and assumes two major data access conditions: logical access (LA) and physical access (PA).
LA covers voice conversion and speech synthesis, while PA covers replay attacks.

The PA task requires the physical process of recording and replaying for data collection and this requirement makes the collection of data for the PA task difficult. 
Therefore, the scale of datasets for replay spoofing detection is relatively limited compared with other domains (e.g. ASVspoof 2019 dataset contains the utterances only from 40 speakers) 
Moreover, the lack of available data also caused the difficulty in generalization to unseen conditions, specifically, channel mismatch conditions \cite{patil2018survey}.
This is considered as an important problem to be solved concerning replay spoofing detection.

To overcome this problem, Shim et. el. \cite{Shim2018ReplaySD} investigated the possibility to use replay configurations as additional information by maximizing the utilization of the available data.
Replay configurations, including the playback and replay devices, as well as the environment, are unique features that exist only in the replayed speech. 
However, it is difficult to obtain labels for these replay configurations.
In the ASVspoof2017 dataset, each device is explicitly labeled; however, the data in the ASVspoof2019 dataset are labeled according to three levels of the device quality: upper, middle, and lower ones.
Providing abstract labels (information that categorizes multiple information into three levels) for a small amount of data can make generalization more difficult.

\begin{figure*}[ht!]
\begin{center}

    \centering
    \includegraphics[width=\linewidth]{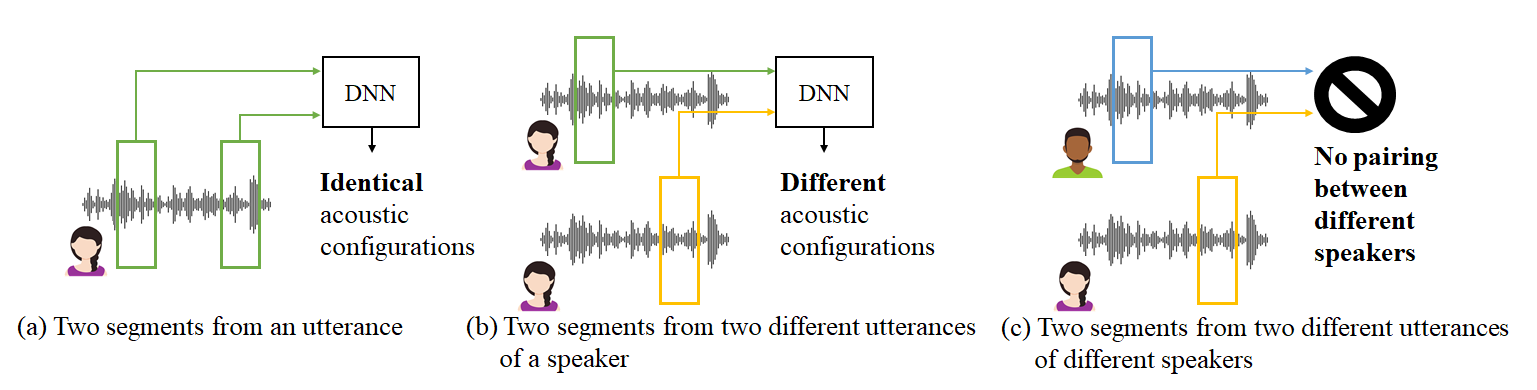}
%\vspace{-1em}
\caption{Self-supervised learning scheme for the acoustic configuration pre-training. The proposed method compares two audio segments to determine whether they have identical acoustic configurations.
We assume that two audio segments have an identical acoustic configuration if they are extracted from the same utterance, or otherwise if they are extracted from different utterances even being produced by the same speaker.}
\label{fig:ss_application}
\end{center}
\vspace{-1.5em}
\end{figure*}

Inspired by this concept, we hypothesize that overall acoustic information can also be used for replay spoofing detection and generalization concerning unseen conditions.
Hereinafter, we define this overall acoustic information as acoustic configurations corresponding to all types of information aside from the voice generated in the process of voice recording, such as microphone types, places, and ambient noise levels.
These are common characteristics inherent in all recorded files, even not replayed ones.
However, they are also difficult to consider, as the labels for acoustic configurations are not available in detail.

In this study, we propose a novel framework utilizing the self-supervised learning scheme \cite{doersch2015unsupervised} that can be used to pre-train for acoustic configurations without a need for explicit labels.
Self-supervised learning is applied to train deep neural networks (DNNs) using data-driven, automatically generated labels instead of the labels manually annotated by humans.
Automatically generated labels indicate naturally available contents, such as the relevant context, correlations, and the embedded metadata.
Self-supervised learning enables pre-training acoustic configurations based on the external data unrelated to the PA task and allows avoiding the problem of overfitting due to the limited PA dataset.
In particular, using the proposed method, it is possible to demonstrate whether the performance can be improved by using additional training data, rather than as proof of how the acoustic configuration can help to detect replay spoofing.
To the best of our knowledge, this is the first approach that applies self-supervised learning to spoofing detection.

We compare two audio segments of the \texttt{YouTube} source data cropped from identical or different utterances, uttered by an identical speaker, and determine whether they belong to an identical utterance.
Our hypothesis is that the two audio segments would have similar acoustic configurations if they are extracted from the same utterance, or otherwise if they are extracted from different utterances, even if they are from the same speaker.
We consider that the proposed method can be utilized to overcome the limitations of available datasets and to improve the generalization performance concerning unseen conditions by learning various acoustic configurations.
Furthermore, by using both \texttt{VoxCeleb1\&2} datasets and intentionally limiting the number of pairs for pre-training, it is confirmed that performance is further improved by pre-training by using a larger amount of data.

\section{Self-supervised learning}

The amount of training data, as well as network capabilities, affects the performance of DNNs.
Although using large-scale datasets with labels helps to achieve better performance, the costs associated with collecting and annotating such datasets are enormous and economically unfeasible.
Concerning the image domain, pre-training based on large-scale datasets, such as ImageNet, and fine-tuning them for other tasks is a commonly used method \cite{girshick2013rich, carreira2017quo, long2015fully}.
The benefit of this approach is that a wide variety of data used in the pre-training phase facilitates better initialization of parameters and enables fast convergence.
Furthermore, when the size of a dataset for the target task is relatively small, learning on the basis of other tasks can allow mitigating the problem of overfitting \cite{jing2019self}.
Therefore, the research has focused on the use of unlabeled data in recent years. 
One of the possible solutions is based on applying self-supervised learning.

Self-supervised learning is a method used to train DNNs with naturally available relevant context obtained from the data without the labels annotated by humans \cite{doersch2015unsupervised}.
The self-supervised scheme is advantageous as it can exploit the unlabeled data as supervisory signals to learn.
Specifically, training in the pretext and downstream tasks can be interpreted as pre-training and fine-tuning for the main-training step, respectively.
Hereinafter in this paper, for consistency, training a pretext task and a downstream one are referred to as pre-training and main-training, respectively.
Generally, pre-training applies self-supervised learning to generate useful feature representations as prerequisite knowledge related to the target task.
After pre-training, the learned parameters are transferred to the main-training step to perform fine-tuning \cite{jing2019self}.
Self-supervised learning was applied to context prediction \cite{doersch2015unsupervised}, image transformation \cite{Gidaris2018UnsupervisedRL}, colorization \cite{zhang2016colorful} in images, and predicting the sequence order \cite{misra2016shuffle} in videos.
It was also exploited in a few research in the audio domain.
For example, \cite{baevski2019vqwav2vec, ravanelli2020multitask} applied self-supervised learning to learn the speech representation in speech recognition, and \cite{Stafylakis2019} proposed to extract speaker embeddings in speaker verification.

\section{Proposed method}

In this study, we propose a two-phase framework that first trains a DNN to learn acoustic configurations by utilizing self-supervised learning, and then performs supervised fine-tuning for replay spoofing detection.
The underlying hypothesis is that pre-training based on acoustic configurations can improve the generalization performance of DNN concerning unseen conditions.
Acoustic configurations represent the environment in which the speaker speaks, microphone types, the distance from the microphone, and the ambient noise.
However, it is difficult to obtain labels for such acoustic configurations.

Self-supervised learning not only can reduce the need to specify detailed information manually, but also enables the use of external data without relevant labels associated with replay spoofing detection. 
Figure \ref{fig:ss_application} illustrates the proposed method of pre-training acoustic configuration using self-supervised learning.
We pre-train the DNN using utterances from various individuals, audio devices, and surroundings included in the \texttt{YouTube} source data, specifically, the \texttt{VoxCeleb1\&2} datasets.
We extract a pair of segments from an utterance and use the pair to teach the DNN that the two segments have the same acoustic configuration. We also extract two segments each from two different utterances from a speaker to teach that the segments have different acoustic configurations.
The speaker labels for pairs were automatically derived from the \texttt{YouTube} data using the method described in \cite{Nagrani2017VoxCelebAL}.
Although unlabeled datasets are generally used for pre-training based on self-supervised learning, we compose each pair from the same speaker only to exclude the impact of the speaker information. That is, two utterances from two different speakers are not used for composing a pair.

\begin{figure*}[!t]
      \centering
      \includegraphics[width=0.7\linewidth]{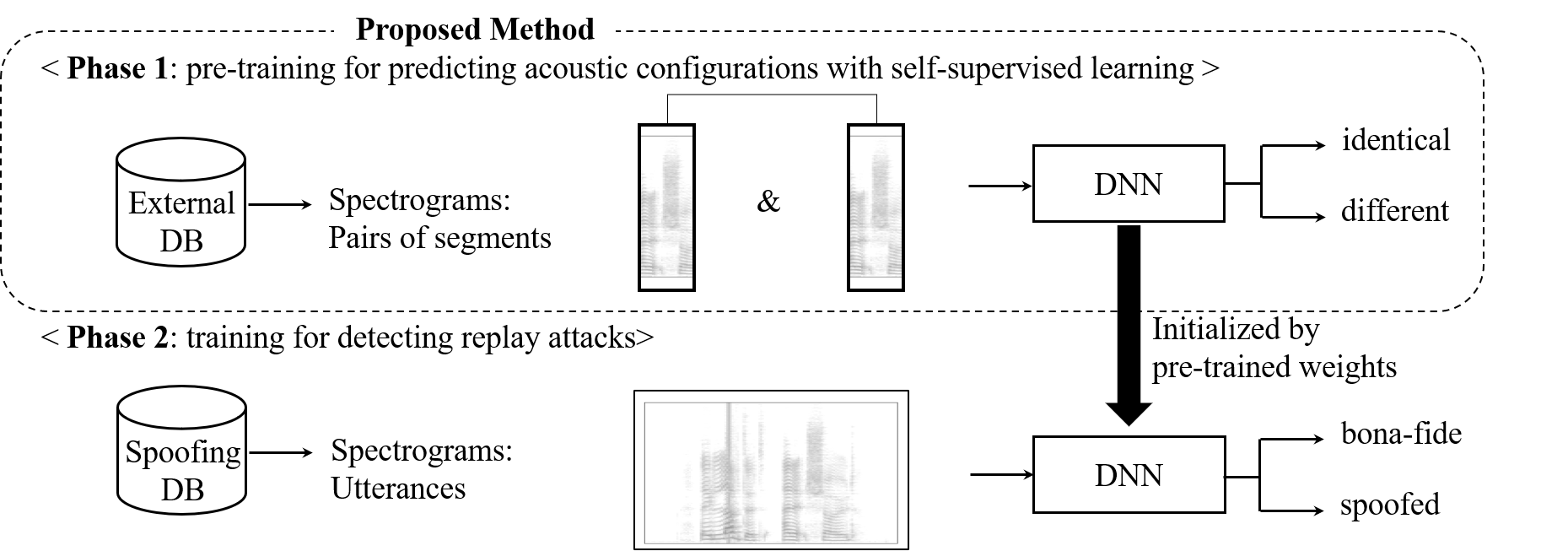}
\caption{Pipeline of the proposed system. There are two phases: self-supervised pre-training of acoustic configurations, and supervised main-training for replay spoofing detection. After the pre-training, knowledge transfers to the main-training with parameters learned from the pre-trained acoustic configurations.}
\label{fig:overall_pipeline}
\vspace{-1.5em}
\end{figure*}

To compare the two segments, we use the cosine similarity.
The loss function is presented in Equation (1), where $x_1$ and $x_2$ are the embeddings from randomly selected audio segments; $\cos(\cdot, \cdot)$ is the cosine similarity between two vectors; and $y$ is the corresponding label of a pair $x$.
If two segments correspond to the same utterance, the label is set to 1; otherwise, the label is set to $-$1.

\begin{equation}
\text{loss}(x, y) =
           \begin{cases}
           1 - \cos(x_1, x_2), & \text{if } y == 1 \\
           \max(0, \cos(x_1, x_2)), & \text{if } y == -1
\end{cases}
\end{equation}

Figure \ref{fig:overall_pipeline} shows the outline of the proposed system.
First, the DNN is pre-trained to differentiate between acoustic configurations in phase 1.
The pairs of audio segments are fed to the DNN and it determines whether two segments in each pair are similar using the cosine distance.
Here, we consider the two segments as similar when they are originated from a single utterance, or otherwise when they are from different utterances.
After the pre-training, we use the pre-trained weights to initializes the DNN for replay detection which is the final goal.
We follow a general scheme of training a replay spoofing detection system as a binary classification task with two categories: bona-fide and spoofed (replayed) \cite{sahidullah2019introduction}.

\section{Experimental settings}
\subsubsection{Dataset}

We used the \texttt{VoxCeleb1\&2} datasets  \cite{Nagrani2017VoxCelebAL, Chung2018VoxCeleb2DS} to implement the proposed pre-training approach
Both datasets consisted of the utterances cropped from \texttt{YouTube} videos that were originally designed for text-independent speaker verification.
\texttt{VoxCeleb1} included 153,516 utterances of 1,251 speakers derived from 22,496 video clips, and \texttt{VoxCeleb2} consist of 1,128,246 utterances of 6,112 speakers fetched from 150,480 video clips.
We used the entire \texttt{VoxCeleb1\&2} datasets for pre-training and also used only a half of \texttt{VoxCeleb1} in the experiment aimed to compare the effect of the number of utterances on performance.
For detecting replay spoofing, we employed the ASVspoof 2019 PA dataset \cite{Todisco2019},
which included 54,000 (5,400 bona-fide, 48,600 spoofed), 29,700 (5,400 bona-fide, 24,300 spoofed), and 137,457 (18,090 bona-fide, 119,367 spoofed) utterances for training, development, and evaluation, respectively.

\subsection{Experimental configuration}

We extracted 2,048 points of magnitude spectrograms for all datasets.
The window length and shift size were 50 ms and 30 ms, respectively.
Utterances of varying duration were used for the test phase for both pre-training and main-training. 
In the training phases, they were cropped into lengths of 200 and 120 frames for the pre-training and the main-training, respectively.
We implemented the system using Pytorch, a deep learning library \cite{paszke2017automatic}.
We utilized the modified end-to-end (E2E) DNN architecture that demonstrated comparable results in the ASVspoof2019 PA condition \cite{Jung2019}.
In the experiments, we employed only a convolutional neural network (CNN) without gated recurrent units layers to perform a simple comparison between the conventional replay spoofing detection scheme and the proposed self-supervised scheme.

\begin{table}[!b]
\caption{The DNN architecture. The numbers in the Output shape column refer to the frame (time), frequency, and number of filters. Conv, BN, and FC indicate convolutional layer, batch normalization, and fully-connected layer, respectively.} % title of Table
\vspace{-1em}
\centering % used for centering table
\begin{tabular}{c c c} % centered columns (4 columns)
\hline %inserts double horizontal lines
Layer & Type & Output shape \\ [0.5ex] % inserts table
%heading
\hline % inserts single horizontal line
\multirow{2}{*}{Conv1} & Conv & \\
 & BN & (120, 1025, 16) \\
\hline
Res block & 
$\left \{
  \begin{tabular}{c}
  BN \\
  %LReLU\\
  Conv\\
  BN\\
  %LReLU\\
  Conv\\
  \end{tabular}
\right \}$

$\times$5

& (15, 17, 128)\\
\hline
MaxPool & Pool & (1, 1, 128)\\
\hline
AvgPool & Pool & (1, 1, 128)\\
\hline
Dense & FC & (64,)\\
\hline
Output & FC & (2,)\\
\hline %inserts single line
\end{tabular}
\label{tab:arc} % is used to refer this table in the text
\end{table}

The E2E DNN included the same residual blocks as in the study by He et al. \cite{he2016identity}, which includes convolutional layers and batch normalization \cite{Ioffe:2015:BNA:3045118.3045167} layers.
Leaky rectified linear unit activation functions \cite{maas2013rectifier} were utilized after each batch normalization.
Table \ref{tab:arc} represents the overall architecture of the E2E CNN.
For further comparison, we considered light convolutional neural networks (LCNNs) \cite{wu2015Light}, which achieved the second-best performance \cite{lavrentyeva2019stc} in the ASVspoof2019 challenge.
LCNNs were also adopted by the winning system in the ASVspoof2017 challenge.
We applied batch normalization after the max feature map and utilized angular margin-based softmax activation \cite{liu2017sphereface} in the end as in \cite{lavrentyeva2019stc}.
The details of the system configurations can be found in \cite{lavrentyeva2019stc}.
We refer to each system as specCNNs and specLCNNs. We use the Adam \cite{Kingma2014AdamAM} optimizer to train both systems.

In particular, in order to confirm the effect of the amount of data used for pre-training on the performance, the experiments were conducted by intentionally limiting the number of pairs.
We initially used 100 pairs per speaker to use a similar amount with the training set configuration of the \texttt{VoxCeleb2} dataset which has 1,128,246 utterances. 
Considering that there are 6,112 speakers in \texttt{VoxCeleb2}, a total of 611,200 pairs per speaker can be used, which have 1,222,400 utterances.
Then we also used 200 pairs per speaker to compare the performance with more data utilization.
The set of pairs were composed of 50\% target and 50\% non-target to balance. 

\begin{table}[!t]
\caption {Experimental results to find the optimal learning rate and batch size. Pre\_lr and main\_lr refer to the learning rates of pre-training and main-training, respectively. The batch size refers to that of the main-training, and the pre-training batch size was fixed to 16.
dev and eval refer to the performance of the development and evaluation sets. }
\label{table:baseline_vox1_result}
    \centering
    \makebox[\linewidth][c]{
    \setlength{\tabcolsep}{6pt}
    \begin{tabular}{lll cc cc}
\hline
 & & & 
\multicolumn{4}{c}{batch size} \\
\cline{4-7}
 & & &  
\multicolumn{2}{c}{32} & \multicolumn{2}{c}{16} \\
\cline{4-5}
\cline{6-7}
system & pre\_lr & lr & dev & eval & dev & eval \\

\hline
\multirow{2}{*}{baseline} & 
\multirow{2}{*}{-} 
& 1 e-4 & 2.91 & 6.89 & 3.22 & 6.60 \\
& & 5 e-4 & 4.24 & 6.88 & 4.29 & 6.36 \\
\hline
\multirow{6}{*}{\texttt{VoxCeleb1}} & 
\multirow{2}{*}{1 e-4} & 1 e-4 & 2.69 & 5.51 & 2.75 & \textbf{5.38} \\
& & 5 e-4 & 2.50 & \textbf{5.15} & 3.44 & 5.85 \\
\cline{2-7}
&
\multirow{2}{*}{5 e-4} & 1 e-4 & 4.90 & 6.00 & 5.87 & 7.0\\
& & 5 e-4 & 3.83 & 5.18 & 6.68 & 7.79 \\
\cline{2-7}
&
\multirow{2}{*}{1 e-3} & 1 e-4 & 6.27 & 9.38 & 5.87 & 8.85 \\
& & 5 e-4 & 4.98 & 7.24 & 6.75 & 7.41 \\
\hline
\end{tabular}
}
\end{table}

\begin{table}[!t]
\caption{Performances depending on the scale of pre-training datasets. “half of \texttt{VoxCeleb1}” comprises only half speakers from \texttt{VoxCeleb1}. POI: Person of Interest.}  % title of Table
\centering % used for centering table
\begin{tabular}{c c c c c} % centered columns (4 columns)
\hline %inserts double horizontal lines
Dataset & 
\multirow{2}{*}{pre\_lr} & 
\multirow{2}{*}{main\_lr} & 
\multirow{2}{*}{dev} & 
\multirow{2}{*}{eval} \\
(\# of POIs) & & & & \\
\hline 
{Half of} & 
\multirow{3}{*}{1 e-4}& \multirow{2}{*}{1 e-4} & \multirow{2}{*}{3.22} & \multirow{2}{*}{6.94} \\ 
\texttt{VoxCeleb1} &  & \multirow{2}{*}{5 e-4} & \multirow{2}{*}{4.03} & \multirow{2}{*}{\textbf{6.66}} \\
(625) & & & & \\
\hline
{\texttt{VoxCeleb1}} & 
\multirow{2}{*}{1 e-4} & 1 e-4 & 2.69 & 5.51 \\
(1,251) & & 5 e-4 & 2.5 & \textbf{5.15} \\
\hline
{\texttt{VoxCeleb2}} & 
\multirow{2}{*}{1 e-4} & 1 e-4 & 2.74 & 4.66 \\
(6,112) &  & 5 e-4 & 2.44 & \textbf{4.64} \\
\hline 
\end{tabular}
\label{table:data} 
\vspace{-1.5em}
\end{table}

\section{Results and analysis}

All experimental results are reported in terms of equal error rates (EERs).
Table \ref{table:baseline_vox1_result} represents the results of comparison between the baseline and proposed systems, using the \texttt{VoxCeleb1} dataset for pre-training.
Considering that the importance of setting an appropriate batch size and a learning rate is emphasized when using a pre-trained model, we investigated the learning rates and batch sizes.
We observed that a smaller learning rate in pre-training and a larger batch size in main-training led to improvements in the performance.
As a result of the experiment, it can be seen that the batch size equal to 32 was better than 16.
We selected the pre-trained model with the best validation performance in terms of spoofing detection.
Pre-training was performed up to a total of 100 epochs, and we observed that the best validation performance was obtained after training around 16–17 epochs in most cases.

As the proposed method can be applied to solve the problem of insufficient data, we attempted to prove whether the proposed study is effective by comparing the performance when adjusting the amount of the pre-training data. That is, the larger is the dataset used for pre-training, the better the performance was expected.
To control the amount of the pre-training data, we investigate the two aspects as follows.
First, to compare the performance according to the scale of the data, we conduct the experiments using half of \texttt{VoxCeleb1} and \texttt{VoxCeleb2} each.
Next, we modify the experimental setup by adjusting the number of pairs within the same dataset.

Table \ref{table:data} represents the results obtained using the different amount of the data for the pre-training phase to evaluate whether enlarging the pre-training dataset is effective.
The results indicated that an increase in the number of utterances in the pre-training phase led to improvements in performance.
On this basis, we assume that the proposed method can be further expanded by considering diverse datasets for various related tasks.
Moreover, we compared freezing points in which we fixed the weights up to a certain layer and did not train them.
However, it was observed that freezing layers resulted in deteriorating performance in all cases.
Therefore, pre-trained parameters were used only for initialization purposes.

Table \ref{table:architecture} demonstrates the robustness of the proposed pre-training approach by employing  LCNN, which is a well-known method for replay spoofing detection.
The obtained results indicated that the proposed approach was also effective concerning LCNN in which we observed approximately a 10 \% relative improvement.
Additionally, it was proved that the performance of the evaluation set improved as the amount of the pre-training data, resulting in EER 3.62\%, 3.57\%, and 3.18\% for a half and whole \texttt{VoxCeleb1} and \texttt{VoxCeleb2} datasets, respectively.

\begin{table}[!t]
\caption{Comparison of system architectures. The same structure was used for pre-training and main-training. That is, if CNN was used for pre-training, CNN was also used for main-training.} 
\centering % used for centering table
\begin{tabular}{c c c} % centered columns (4 columns)
\hline %inserts double horizontal lines
System & 
w/o pre-training & w/ pre-training \\
\hline 
{specCNNs} & 6.36& \textbf{4.64} \\ 
{specLCNNs} & 3.50 & \textbf{3.18}\\
\hline 
\end{tabular}
\label{table:architecture} 
\end{table}

Finally, we conducted an additional experiment to verify the validity of the proposed approach by doubling the number of pairs while using the same dataset. 
To achieve this, we designed the experiments to use the limited number of pairs and to adjust it instead of using as many pairs as possible by randomly selecting pairs per batch.
Table \ref{table:pair} shows the results.
Here, the hypothesis was that if the proposed approach is effective, solely doubling the number of pairs would also allow improving the performance.
At this time, we used LCNN which showed superior performance in the previous experiment. 
The results showed that the performance improved with an increase in the number of pairs.
%Based on these results, we consider that the proposed framework can further leverage other large-scale datasets for the related tasks such as speech recognition, and thereby achieve further improvements.

\begin{table}[!t]
\caption{The effect of increasing the number of pairs. By intentionally limiting the number of pairs, it can be demonstrated the effect of the amount of data used for pre-training, while the dataset is fixed.}
\centering % used for centering table
\begin{tabular}{c c c c} % centered columns (4 columns)
\hline %inserts double horizontal lines
\multirow{2}{*}{System} & 
w/o & w/\\
{} & doubling pairs & doubling pairs \\
\hline 
specLCNNs & 3.18 & \textbf{2.65} \\ 
\hline 
\end{tabular}
\label{table:pair} 
\vspace{-1.5em}
\end{table}

\section{Conclusion}

In this study, we proposed a pre-training scheme based on self-supervised learning for replay spoofing detection.
To overcome the problem of the limited amount of data available for replay spoofing detection, we assumed that training on acoustic configurations derived from the audio datasets unrelated to replay spoofing could improve the performance.
Therefore, we applied self-supervised learning for the training of acoustic configurations.
The experiments are conducted using the datasets of ASVspoof2019 and \texttt{VoxCeleb1\&2} for replay spoofing detection and pre-training acoustic configurations, respectively. 
%As a result, we achieved performance improvement over 30\% in EERs compared to the baseline system.
As a result, the proposed system reduced the EER from 6.36\% to 4.64\% with the same network of specCNNs, and further improvement was achieved by adjusting the network architecture and the number of pairs.
%The proposed method not only proposed a way to utilize external data from other domains for replay attack detection but also can be used in various ways in the future because it can utilize data without annotation.

\section{Acknowledgements}
This research was supported by Projects for Research and Development of Police science and Technology under Center for Research and Development of Police science and Technology and Korean National Police Agency funded by the Ministry of Science, ICT and Future Planning (Grant No. PA-J000001-2017-101)

\clearpage
\bibliographystyle{IEEEtran}
\bibliography{ref}

\end{document}